\documentclass[10pt, a4paper]{article}

\usepackage[final]{lrec2026} 

\usepackage{listings}
\lstdefinestyle{XMLstyle}{language=XML}
\newcommand{\xml}[1]{\lstinline[style=XMLstyle]{#1}}

\title{MioFFAn: an Annotation Software for Formula Formalization with LLM Automation Capabilities}

\name{Nicolas Sibuet$^{*1}$, Horacio Saggion$^{2}$, Riccardo Rossi$^{1,3}$} 

\address{$^{1}$Universitat Politècnica de Catalunya (UPC), Barcelona, Spain\\
$^{2}$Universitat Pompeu Fabra (UPF), Barcelona, Spain\\
$^{3}$International Center for Numerical Methods in Engineering (CIMNE), Barcelona, Spain. \\
\{nicolas.sibuet, riccardo.rossi\}@upc.edu, horacio.saggion@upf.edu\\}

\abstract{
The automatic translation of mathematical expressions in scientific literature into executable symbolic code—a process we refer to as Formula Formalization—is hindered by a severe scarcity of high-quality, ground-truth datasets specialized for technical scientific domains. In this paper, we present MioFFAn, an open-source, document-centric, and customizable framework designed to facilitate rapid annotation for this task. Building upon the MioGatto architecture, we extend existing features to overcome structural limitations and pivot its scope by introducing specific functionalities for Formula Formalization, such as selection of equations of interest and aided symbolic code specification. By allowing users to configure custom taxonomies and properties for identified symbols, and compatible symbolic operators, we ensure the framework is adaptable to diverse specialized scientific fields. Furthermore, MioFFAn is designed to incorporate partial automation via Large Language Models. By defining a modular set of automated sub-tasks with strict output formats, we enable researchers to iteratively refine automation capabilities and evaluate competing strategies using standard NLP metrics. We specify the current automation methodology and perform a preliminary evaluation that demonstrates to efficacy of this human-in-the-loop approach.
 \\ \newline  }

\begin{document}

\maketitleabstract

\section{Introduction}
Scientific literature serves as the primary vehicle for communicating complex discoveries across all disciplines. For many fields, these discoveries are expressed through mathematical formulations that are to be implemented computationally. This implementation process requires both precise semantic interpretation of mathematical notation and the technical ability to manipulate these elements within a programming language. While modern Optical Character Recognition (OCR) systems effectively extract presentational mathematical formats—such as LaTeX or MathML—from PDF documents, these lack the semantic and computational depth required for the extracted mathematical formulas to be implemented in code. As noted by \citet{greiner-petter_making_2023}, converting these representations into unambiguous, computationally complete models would require systems capable of synthesizing relevant contextual information, such as explicit mentions of variables or operators and their case-specific properties. Although Large Language Models (LLMs) offer a promising mechanism for orchestrating this transformation, their progress is hindered by the scarcity of high-quality, case-specific datasets that can serve as benchmark or training corpora.

In this work, we present MioFFAn (Math identifier-oriented Formula Formalization
Annotator)\footnote{MioFFAn is open-source and available at \url{https://github.com/NicolasSR/MioFFAn}}, an annotation tool designed for the fine-grained labeling of mathematical expressions within scientific documents to establish the ground-truth symbolic code required to model them in a Computer Algebra System (CAS). We hereafter refer to this task as Formula Formalization. We build upon the MioGatto framework \cite{asakura_miogatto_2021}, pivoting from its original scope to introduce features tailored to our requirements. 
In particular, the original MioGatto software was designed for description alignment \cite{alexeeva_mathalign_2020} and coreference analysis of mathematical entities, focusing on the evolution of local notations throughout a document. Therefore, it addresses individual symbols rather than the way they are combined within formulas. In contrast, MioFFAn encompasses symbol characterization and grounding but extends to the generation of symbolic code that orchestrates these elements.

Our software provides domain-specific flexibility, allowing users to customize annotation parameters according to the unique constraints of different scientific fields, such as physics, economics and biological modeling, among others. The system is document-centric and interactive, enabling users to maintain full context by annotating directly onto the manuscript. Finally, the framework is designed for semi-automation in a modular way: by specifying certain tasks to be done automatically and providing corresponding interfaces and evaluation routines, developers may implement automation strategies in any way they see fit. While we incorporate baseline automation via LLM in this work, the system is designed to facilitate more sophisticated automated workflows in the future.

To demonstrate the applicability of our framework in realistic research and development scenarios, we present a use case centered on Finite Element Methods (FEM) \cite{zienkiewicz_finite_2024}, cornerstone of numerical simulation. We illustrate how MioFFAn can be configured to characterize this domain's variable properties and operators and we use this use case within a proof-of-concept evaluation of current automation capabilities.

The remainder of this paper is organized as follows: Section \ref{sec:related_work} reviews related work in terms of similar NLP tasks, existing datasets and prior annotation tools. Section \ref{sec:the_mioffan_framework} thoroughly describes the MioFFAn framework, including a small review of the original MioGatto tool. Then, Section \ref{sec:use_case} describes the use case to which MioFFAn has been configured during this study. Following, Section \ref{sec:proof_of_concept_evaluation} details the proof-of-concept evaluation for the current automation framework. Section \ref{sec:mioffan_in_external_pipelines} discusses possible integration of MioFFAn into other pipelines and Section \ref{sec:limitations} specifies the most important limitations of the software. Finally, Section \ref{sec:conclusions_and_future_work} concludes the document and proposes future work directions.

\section{Related work}
\label{sec:related_work}
\subsection{Related NLP tasks}
We review the tasks that relate the most to Formula Formalization:

\textbf{Neural machine translation (NMT) for mathematics}. The translation of mathematical notation from presentational formats (e.g., LaTeX) to computational formats can be framed as a particular NMT \cite{kalchbrenner_recurrent_2013, tan_neural_2020} task. \citet{petersen_neural_2023} use this terminology and demonstrate that end-to-end NMT can outperform rule-based systems in LaTeX-to-Mathematica translation; however, they highlight a significant bottleneck for further generalization caused by notation bias, lack of contextual integration and insufficient hand-crafted data. Prior rule-based pipeline LaCaST \cite{greiner-petter_math_2023} leverages external convention databases and entity linking to address ambiguity, though it lacks the flexibility of deep learning architectures.

\textbf{Tasks on mathematical grounding and tagging}. Relating specific mathematical identifiers with their in-document descriptions is attempted via LLM in \cite{dev_approach_2024} but also before, via rule-based systems \cite{schubotz_semantification_2016, alexeeva_mathalign_2020}. Mathematical Entity Linking \cite{kristianto_entity_2016} strives to associate mathematical symbols to knowledge base entries corresponding to the concept they represent. Finally, Part-of-Math (PoM) Tagging \cite{youssef_part--math_2017} classifies mathematical symbols according to the type of concept that  they represent, according to a specific taxonomy. \citet{zou_stem-pom_2025} and \citet{shan_using_2024} approach the task by using LLMs and determine that further work is needed to recognize complex math concepts and that explicit human expert evaluation is needed for proper assessment.

\textbf{Autoformalization}. The Autoformalization task \cite{weng_autoformalization_2025} refers to the translation of natural language mathematics into verifiable formal languages (e.g., Lean, Isabelle). \citet{zhang_autoformalization_2025} conclude that including well-defined contextual information and syntactic correction capabilities are determinant to achieve good results. While the task is generally focused on formal proofs, recent expansions have applied these techniques to physics lemmas and theorems \cite{soroco_pde-controller_2025, kabra_can_2025}.

\textbf{Word math problem solving}. Finally, research into solving word math problems has shown that offloading reasoning to symbolic solvers or CAS significantly improves accuracy \cite{he-yueya_solving_2023, chen_program_2023}. These systems generate intermediate symbolic programs—comprising characterized variables and equations—that align closely with our proposed Formula Formalization output, despite differing in input modality and end goals.

\subsection{Related Datasets}

Ground-truth data related to Formula Formalization is notably scarce. MathMLben \cite{schubotz_improving_2018} provides \textasciitilde300 LaTeX expressions in tree representation with Wikidata-linked variables, while MathAlign \cite{alexeeva_mathalign_2020} links identifiers to textual descriptions in arXiv snippets. However, these resources often omit complex expressions and lack the domain specificity required for real-world applications. Recent benchmark STEM-PoM \cite{zou_stem-pom_2025} focuses exclusively on the PoM Tagging task. In contrast, Autoformalization benchmarks (e.g., MiniF2F \cite{zheng_minif2f_2022}, LeanEuclid \cite{murphy_autoformalizing_2024}) are highly structured but strictly limited to proof assistants, failing to support the general-purpose CAS integration or numerical simulations required for engineering contexts.
\citet{schembera_towards_2025} present a graph knowledge base for applied mathematics and algorithms. It incorporates mathematical models and formulations for several scientific fields, indicating the nature of the quantities included in each formulation, but does not orchestrate corresponding symbolic codes or semantic trees. The knowledge base is collaborative, but limited by time-consuming manual input.

\subsection{Mathematical Annotation}

Existing frameworks for mathematical annotation vary significantly in input modality and granular focus. Early annotation tools compatible with mathematical notation, like KAT \cite{schmoll_kat_2016} and AnnoMathTeX \cite{scharpf_annomathtex_2019} leverage HTML5 and LaTeX, respectively. The latter incorporates recommendation systems from Wikipedia and arXiv. For PDF-based workflows, \citet{alexeeva_mathalign_2020} provide a specialized identifier annotator. MioGatto \cite{asakura_miogatto_2021} and AnnoTize \cite{panzer_annotize_2023} both utilize MathML in HTML environments; while MioGatto emphasizes rapid annotation through mathematical identifier recognition and allows for contextual coreference resolution via text highlighting, AnnoTize prioritizes extensive user personalization for annotated object properties and marking style. VARAT \cite{kato_varat_2025} initiates a movement toward domain-specific workflows. It introduces industry-specific annotation for chemical manufacturing, although the annotation content is limited to textual definition and grounding on whole paragraphs. Unfortunately, none of these tools implement specific features for annotating the symbolic code representation of whole mathematical expressions.

The MathMLben suite \cite{schubotz_improving_2018} focuses on graph representations of complete equations and the STEM-PoM Labeler \cite{zou_stem-pom_2025} focuses on taxonomic symbol classification. However, these frameworks process isolated equations, losing the document-level context essential for disambiguation.

Critically, while MioGatto and the STEM-PoM Labeler have been used in research involving LLM-assisted automation \cite{zou_stem-pom_2025, dev_approach_2024}, the implementation of the frameworks used for LLM interaction are not publicly available.

\section{The MioFFAn Framework}
\label{sec:the_mioffan_framework}

Designed as a document-centric suite utilizing HTML and MathML, the framework facilitates a multi-stage annotation pipeline that incrementally reduces the complexity of Formula Formalization. The main steps in this pipeline are:
\begin{enumerate}
    \item Formula Selection: The user identifies a target mathematical expression within the manuscript.
    \item Identifier Characterization: The user defines mathematical properties and attributes for the specific identifiers within that expression.
    \item Contextual Grounding: These specifications are grounded by aligning identifiers with relevant context segments through document-level highlighting.
    \item Symbolic Synthesis: The user writes the symbolic code for the formula, leveraging the previously grounded concepts and CAS operators.
\end{enumerate}

MioFFAn utilizes a client-server model (TypeScript/Python) and operates locally via web browser. The choice of HTML+MathML over LaTeX, which is the preferred markup language for writing scientific corpora, is that they are specifically designed to be integrated into dynamic web applications, facilitating selection of elements and transformations of these via well-established web-app development suites. Nonetheless, conversion from LaTeX files to HTML and vice-versa may be added to the software at a later stage. Furthermore, there is increasing interest from scientific paper repositories such as arXiv towards using HTML as a more accessible visualization format \cite{arxiv_accessible_nodate}.

\begin{figure}[!ht]
\begin{center}
\framebox{\includegraphics[]{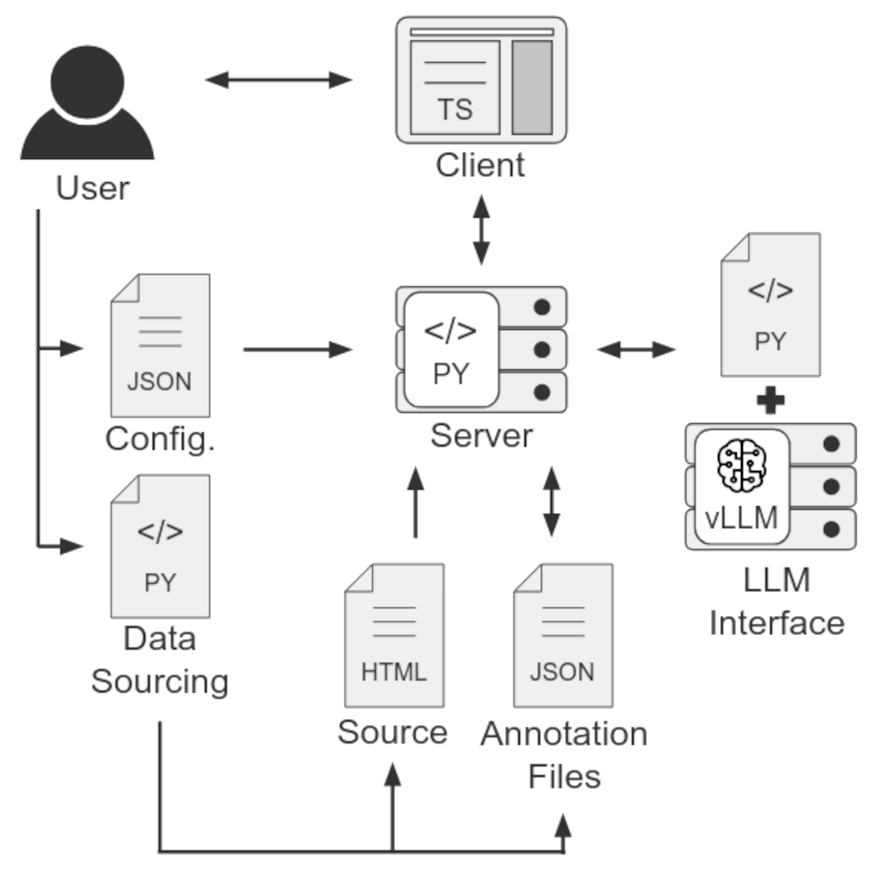}}
\caption{Architecture schematic of MioFFAn}
\label{fig:schematic}
\end{center}
\end{figure}

\begin{figure}[!ht]
\begin{center}
\framebox{\includegraphics[]{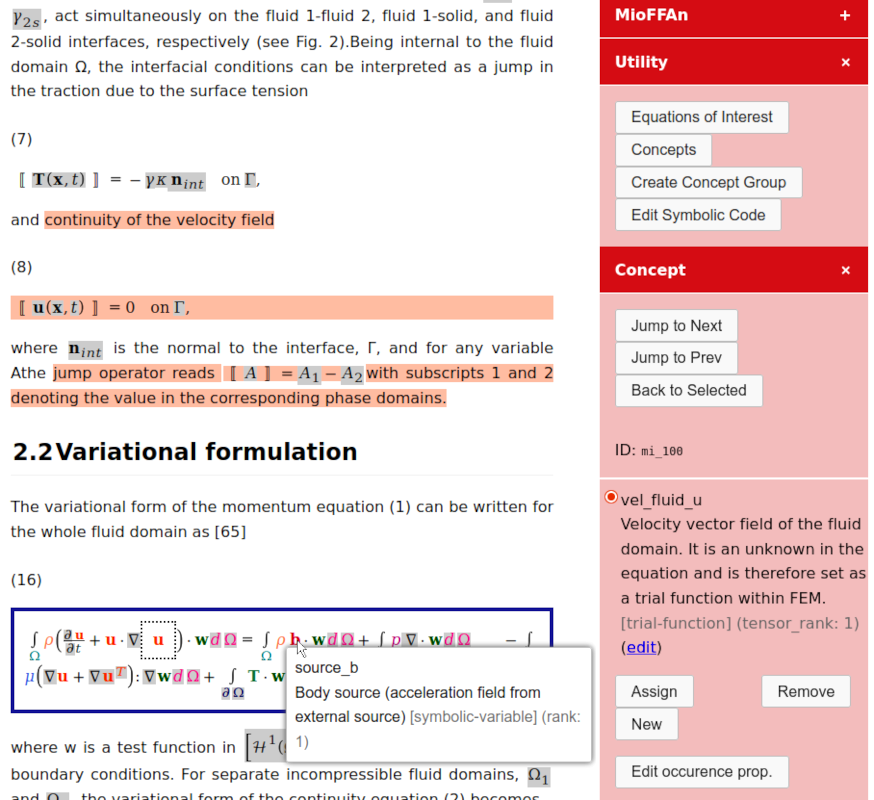}}
\caption{MioFFAn's interface with annotations. The EoI is surrounded by a blue frame. Colored symbols indicate Occurrences. SoGs are highlighted (orange) for the selected Occurrence ($\mathbf{u}$). Information is shown for the Occurrence under the mouse pointer ($\mathbf{b}$). Edition toolbar on the right, in red. The sample content is sourced from \cite{hashemi_three_2021}, as is in all subsequent figures.}
\label{fig:global_view}
\end{center}
\end{figure}

A schematic of MioFFAn's architecture can be viewed in Figure \ref{fig:schematic}, while a global view of the software's interface can be observed in Figure \ref{fig:global_view}. The content of the sample shown in all the figures that contain MioFFAn's interface is sourced from \cite{hashemi_three_2021}.

As the framework builds upon the MioGatto architecture \cite{asakura_miogatto_2021}, we first provide a preliminary analysis of the base software's characteristics before detailing our specific extensions and architectural shifts.

\subsection{Description of the Original MioGatto Software}
\label{ssec:study_of_original_miogatto_software}

We selected MioGatto as our foundational framework due to its MathML-native architecture, its document-centric approach, its interactivity and its approach to fast identifier selection. Most importantly, its features align with the identifier characterization and contextual grounding steps from our proposed pipeline at the beginning of Section \ref{sec:the_mioffan_framework}.

The system implements three core functionalities:
\begin{itemize}
    \item Math Identifier (MI) Management: Automated identification of <mi> MathML tags, enabling rapid selection via mouse or keyboard.
    \item Concept Assignment: Annotation of a given MI is done by assigning a Concept (description, arity, and notation affixes) to it. An MI associated with a Concept is termed an Occurrence\footnote{Throughout this paper, capitalized \textit{Concept} and \textit{Occurrence} refer specifically to these architectural entities.}; all Occurrences of the same Concept share a distinct color for visual consistency and distinguishability.
    \item Contextual Grounding: Highlighting of text segments to serve as Sources of Grounding (SoG), linking descriptions or relevant information directly to specific Occurrences.
\end{itemize}

Despite its strengths, MioGatto presents several bottlenecks for its effective use in realistic annotation dataset creation. Architecturally, the system is constrained by a one-sample-per-session limit, requiring a server restart to switch documents. Data-wise, its annotation schema pre-allocates entries for every MI regardless of activity. Functionally, we identify four critical limitations: (1) Annotations are strictly limited to single <mi> tags (generally single characters), preventing the grouping of complex symbols; (2) Concept properties are fixed and mix intrinsic characteristics (arity) with notation-specific ones (affixes); (3) SoG highlighting is restricted to individual <span> nodes, preventing cross-node or in-line math selection; and (4) Concepts are identified by the Unicode characters they contain, making the underlying annotation structure cumbersome to manipulate.

\subsection{Enhancement of Original Features}

We implemented architectural enhancements that resolve the limitations identified in Section \ref{ssec:study_of_original_miogatto_software}.

To allow for \textbf{hierarchical and compound MI management}: MioFFAn extends the definition of a Math Identifier beyond single \xml{<mi>} tags. By default, the system now natively supports compound MathML structures, including sub-scripts (\xml{<msub>}), super-scripts (\xml{<msup>}), and their combinations (\xml{<msubsup>}). The user may add more tags as they see fit via JSON configuration. All MIs are uniquely identified by tag IDs.

\begin{figure}[!ht]
\begin{center}
\includegraphics[]{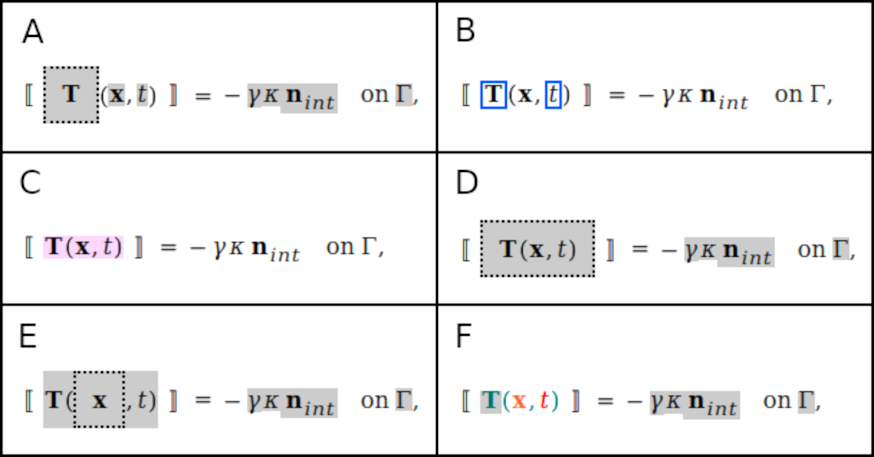}
\caption{Group designation process and example. Sequentially: (A) Annotation options (shadowed symbols) prior to grouping. (B) Within Group creation tool: first and last Group elements selected by mouse click. (C) Within Group creation tool: complete group visualization. (D) Annotation options after grouping. (E) Demonstration of inner MI selection within Group. (F) Example view of hierarchically assigned concepts.}
\label{fig:group_global}
\end{center}
\end{figure}

To capture complex notations like $\delta u_i$ or $f(x)$ as single entities, we introduce Groups: user-defined MIs encompassing multiple contiguous sibling elements within the MathML tree. These are implemented via a temporary \xml{<mstyle class="custom-group">} wrapper in the server-side DOM, preserving the integrity of the original source file. They are permanently and uniquely identified in the annotation files by the first and last \xml{<mi>} tags within their combined sub-trees in Depth-First Search (DFS) traversal order. The declaration process for Groups is intuitive, needing only the selection by mouse of start and end characters to include. An example of the process can be visualized in Figure \ref{fig:group_global}. This way, the user does not need to deal with the explicit HTML code.

This architecture implicitly enables hierarchical annotation, where a Concept can be assigned to a holistic expression (e.g., a function $f(x)$) while its constituent symbols (e.g., $x$) retain their own specific Concept assignments. To navigate overlapping tags, users employ assigned keyboard keys to perform a DFS traversal of the MI tree, ensuring precise selection regardless of visual density.

To \textbf{decouple the property schema}: we allow both Concepts and Occurrences to have their own independent properties. This way, we differentiate:
\begin{itemize}
    \item Invariant Semantic Properties (Concept): Defined once for a mathematical entity (e.g., a "velocity vector" ($\textbf{u}$) is always a vector quantity).
    \item Variant Properties (Occurrence): Specific to a single instance (e.g., whether the vector is currently transposed ($\textbf{u}^T$)).
\end{itemize}

This disentanglement allows the same Concept to be reused across different implementation modalities and notations, instead of needing to create slightly different Concepts for each scenario. Users can customize these property sets via JSON configuration, enabling the tool to adapt to the specific requirements of different scientific disciplines. A view of the interface for defining new Concepts in MioFFAn can be seen in Figure \ref{fig:new_concept}.

To \textbf{improve SoG flexibility}: SoGs now support inter-span highlighting, allowing users to select text across multiple HTML nodes, including in-line mathematical expressions. It also allows for the highlighting of entire displayed formulas. SoGs are now directly associated to Concepts, instead of individual Occurrences.

Additionally, we allow the annotation schema to be dynamically populated as needed and the dependency to Unicode is replaced by unique IDs for Concepts. Finally, users can now navigate between document samples within a single session without requiring a server restart, facilitating large-scale dataset curation.

\subsection{Formula Formalization-Specific Features}

We introduce three core features that are vital to the Formula Formalization workflow.

\begin{figure}[!ht]
\begin{center}
\framebox{\includegraphics[]{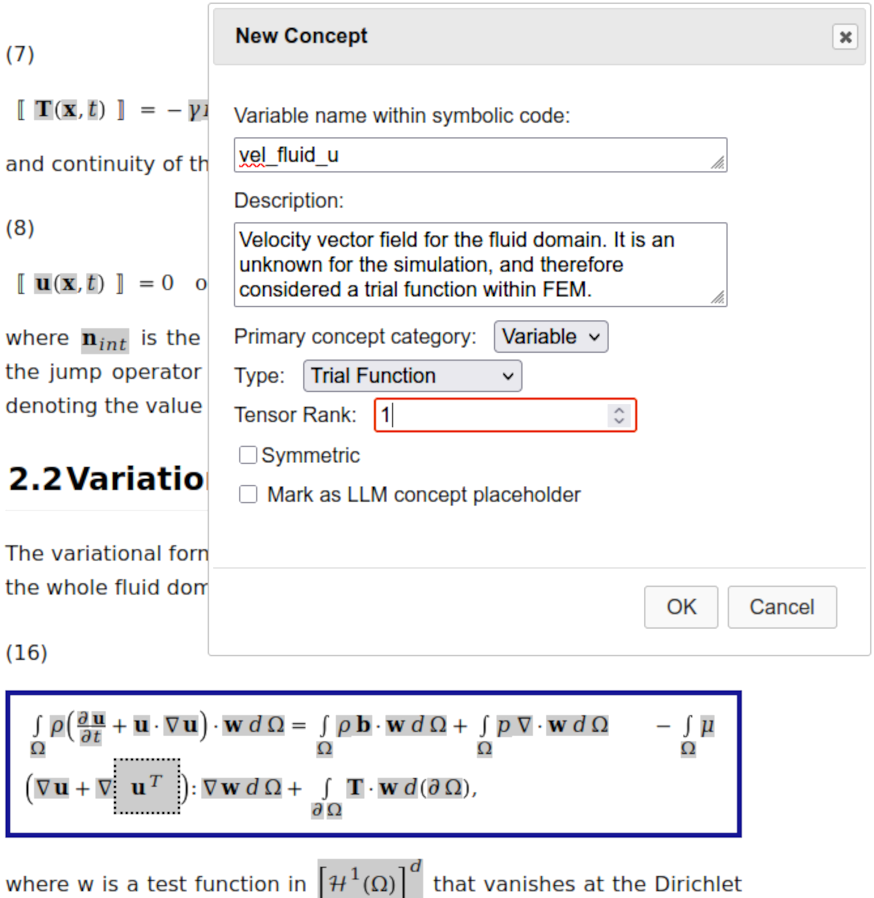}}
\caption{Menu for the registration of a new concept for the selected MI ($\mathbf{u}^T$). Fields according to taxonomy in Section \ref{sec:use_case}. EoI visibly marked by a blue frame.}
\label{fig:new_concept}
\end{center}
\end{figure}

First, we introduce \textbf{Equations of Interest (EoI)}. Rather than treating all document content equally, users designate specific EoIs, susceptible of formalization. This visually marks the expression and initializes a dedicated entry for that formula in the annotation schema. Each EoI is indexed via the HTML ID of its parent <div class="formula"> tag. The view of a marked EoI within the document can be seen within Figures \ref{fig:global_view} and \ref{fig:new_concept}.

Second, we introduce \textbf{canonical Variable Names}. To ensure that defined Concepts can be uniquely represented within the annotated symbolic code, the framework requires each Concept to be mapped to a Variable Name.

\begin{figure}[!ht]
\begin{center}
\framebox{\includegraphics[]{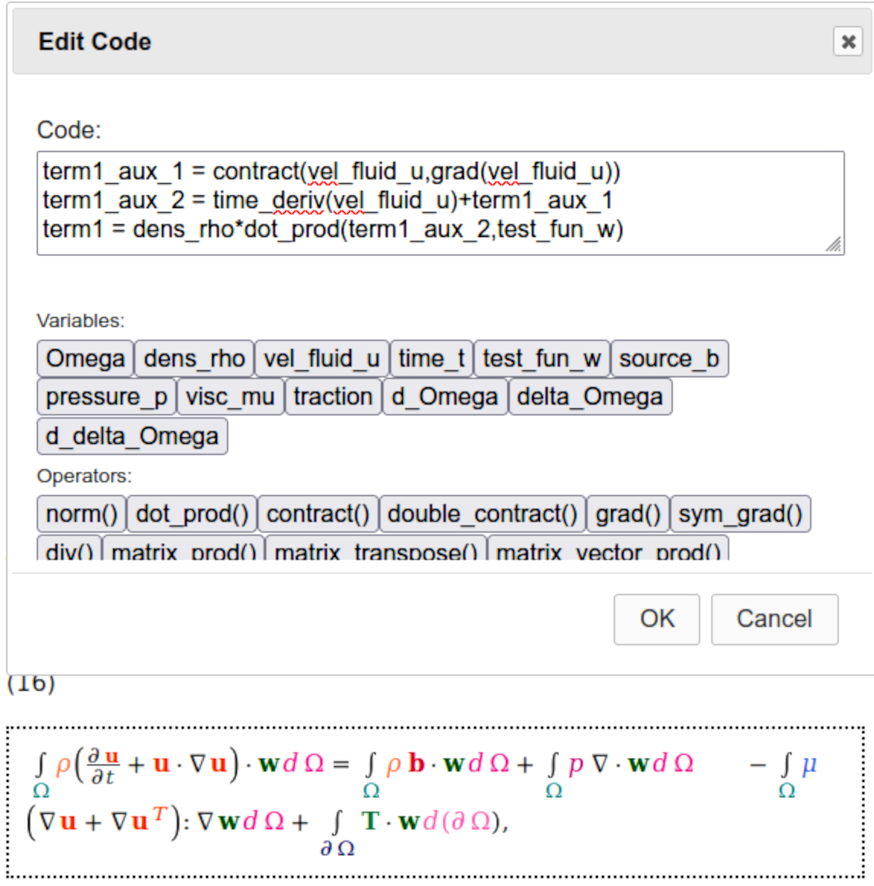}}
\caption{Menu for the specification of the symbolic code for an EoI. Shows palettes for both variables and operators.}
\label{fig:code_editor}
\end{center}
\end{figure}

Third, we introduce \textbf{integrated symbolic code construction}. The final stage of the pipeline is the synthesis of the symbolic code representation. To minimize manual syntax errors and accelerate the translation process, MioFFAn provides an interactive construction environment with a palette containing the Variable Names for all Concepts defined in previous steps and a list of available operators specific to the target CAS. When a user selects an element from these lists, it is automatically injected into the text box for the code. However, these aids do not stop the user from defining their own intermediate variables and logic manually in order to maintain readability or handle complex local logic. A view of this functionality can be seen in Figure \ref{fig:code_editor}.

\subsection{Data Sourcing, Management and Preprocessing}

To ensure high-fidelity structural representation, we implement a preprocessing routine tailored for the ScienceDirect API\footnote{\url{https://dev.elsevier.com/}}, as it supports the retrieval of research papers in XML format, bypassing the need for non-deterministic OCR and instead performing a direct transformation into MathML-compliant HTML.

During this transformation, we perform a deterministic ID injection phase:
\begin{itemize}
    \item  Mathematical Entities: All \xml{<mi>} tags, designated compound MI tags and \xml{<div class="formula">} containers are assigned unique persistent identifiers if they are not already present in the source XML.
    \item Textual Context: To facilitate precise grounding, the text is segmented into \xml{<p>} nodes and \xml{<span>} nodes within them, each receiving a unique ID. 
\end{itemize}

This system allows the annotation schema to reference specific tags in the document. Said annotation system is based on two different JSON files: \textit{anno.json}, used for Groups and \textit{mcdict.json}, used for Concepts, EoIs and Occurrences. Both of them are generated during the preprocessing step.

\subsection{Partial Automation via LLM}

MioFFAn integrates infrastructure for LLM-assisted annotation. Rather than an end-to-end approach, we implement a modular, stage-wise pipeline that allows for human-in-the-loop verification and refinement. The automation logic is implemented in Python and is decoupled from the core application, enabling researchers to swap or evaluate different strategies without modifying the software's primary interface and functionalities.

In terms of interface, we utilize the OpenAI-compatible vLLM API to interact with local Large Language Models. This choice ensures data privacy—a critical requirement in scientific research.

The framework currently supports three automation tasks designed to populate the annotation schema:
\begin{itemize}
    \item \textbf{Symbol Segmentation:} The model identifies which MathML elements or compound structures within an EoI should be grouped as unique Occurrences. For example, in $\delta uF(x)$, the system suggests segments for $\delta u$ and $F(x)$. These are assigned placeholder Concepts to allow for easy visual verification and correction. The software enables the user to assign placeholder Concepts manually as well, in order to annotate ground-truth samples for this specific task, and in order to correct the automated results before continuing to the next steps. These placeholders do not require any description or Concept property specification.
    \item \textbf{Concept Assignment:} The system suggests Concepts, populated with at least their canonical Variable Names and a textual description. Within this same stage, the model maps these Concepts to existing placeholder Concepts.
    \item \textbf{SoG Identification:} The model identifies text spans or formulas in the HTML tree that serve as the SoG for each Concept. By providing the start and end node IDs, the system automatically generates document-level highlights.
\end{itemize}

By providing the HTML tree and the content within the annotation files as input for the routines and  enforcing a fixed output format at each stage (lists of Groups, Concepts, Occurrences or SoGs to be added), MioFFAn enables methodology-agnostic evaluation. Predicted annotation outputs can be directly compared against gold-standard annotation files, and the annotation files can be processed together with the source HTML file to apply standard NLP metrics. The fact that MioFFAn annotations are based on pointers towards content included in the original HTML file, and not the textual content itself, makes it less susceptible to format-related bias during evaluation. We acknowledge that this limits the scores for systems that generate correct natural language content but are not able to locate it correctly within the original document. However, we believe this relation capability to be an integral necessity for the transparency of these systems towards the human annotator and therefore we consider this effect to be fair. In any case, the user may implement intermediate LLM logs within their automation framework, together with custom evaluation metrics, to directly evaluate natural language responses against the gold standard, bypassing this constraint.

\subsubsection{Provisional Automation Methodology}
\label{secsec:provisional_automation_methodology}

The automation code and LLM prompts used for this paper are available within the examples directory of MioFFAn's GitHub repository. Here we describe the process followed for each task. For the \textbf{symbol segmentation}:

\begin{enumerate}
    \item The LLM is tasked with recreating the MathML sub-structures for candidate symbols.
    \item A Python routine using the lxml library performs a search within the source MathML tree to validate these candidates. Proposed structures that lack an exact match in the original source are discarded.
    \item If overlapping structures are detected (e.g., $x$ nested within $F(x)$), a second LLM call provides the conflict context and solves it, either removing the inner structure instance or maintaining it as an independent symbol.
\end{enumerate}

For the \textbf{concepts assignment}:

\begin{enumerate}
    \item The LLM proposes Concepts relevant to the EoI. For each one, it defines a unique Variable Name, a textual description, and a Justification for its choice (appended to the description).
    \item The LLM performs a mapping from these new Concepts to the available segmented symbols. Concepts that fail to find a specific symbol occurrence are pruned.
    \item For every Concept individually, the LLM determines appropriate domain-specific properties.
\end{enumerate}

For the \textbf{SoG identification}:
\begin{enumerate}
    \item For each Concept, the LLM identifies the IDs of specific \xml{<span>}, \xml{<p>} or \xml{<div calss="formula">} tags that serve as textual evidence for the Concept's meaning and properties.
    \item The framework verifies that the returned IDs exist within the document before finalizing the SoG highlights in the annotation file.
\end{enumerate}

\section{Use Case}
\label{sec:use_case}

\begin{table*}[t] 
\centering
\begin{tabularx}{\textwidth}{|l|l|X|X|}
      \hline
      \textbf{Category} & \textbf{Level} & \textbf{Categorical Properties} & \textbf{Single-Value Properties}\\
      \hline
      Variable & Concept & \textbf{Type}: [Trial, Test, Nodal, Symbolic, Numerical] & Tensor Rank, Symmetry \\
               & Occurrence & --- & Transpose, Voigt \\
      \hline
      Function & Concept & \textbf{Type}: [Defined, Undefined] & Tensor Rank, Symmetry, Dependencies \\
               & Occurrence & --- & Transpose, Voigt \\
      \hline
      Operator & Concept & \textbf{Type}: [Differential, Tensorial, Algebraic] &  Linearity \\
      \hline
      Domain   & Concept & \textbf{Type}: [Whole, Boundary] & Dimension \\
      \hline
      Integration var. & Occurrence & --- & Domain Name Variable \\
      \hline
\end{tabularx}
\caption{Taxonomy of mathematical concepts and occurrence-specific properties for the FEM use case.}
\label{tab:fem_taxonomy}
\end{table*}

\begin{table*}[!ht]
\begin{center}
\begin{tabular}{l|rr|r|r|}
\cline{2-5}
& \multicolumn{2}{l|}{Symbol Segm.} & Concepts Assign. & SoGs Ident. \\ \cline{2-5} 
& \multicolumn{1}{l|}{Coverage (\%)} & \multicolumn{1}{l|}{CoNLL (\%)} & CoNLL (\%) &  Avg. SDI (\%)               \\ \hline
\multicolumn{1}{|l|}{Explicit Names}      & \multicolumn{1}{r|}{36.2}     & \multicolumn{1}{r|}{26.9}  & 97.0   & 15.3 \\ \hline
\multicolumn{1}{|l|}{Original Characters} & \multicolumn{1}{r|}{57.7}     & \multicolumn{1}{r|}{36.8}  & 96.9  & 14.9 \\ \hline
\end{tabular}
\caption{Average results for each evaluation metric, comparing the unmodified automation workflow (Original Characters) with a modified workflow that substitutes non-standard Unicode characters by their explicit names (Explicit Names).}
\label{tab:llm_results}
 \end{center}
\end{table*}

We configure the framework for its use in Finite Element Methods (FEM), a domain characterized by high-dimensional tensors and variational formulations that demand precise characterization. Specifically, we target a custom Sympy-based symbolic compiler\footnote{\url{https://github.com/NicolasSR/KratosFECompiler}} designed to translate certain variational forms of partial differential equations into simulation-ready C++ code for the KratosMultiphysics \cite{dadvand_object-oriented_2010} framework.

The configuration encompasses, on one side, the \textbf{domain taxonomy and properties schema}: We define a tiered schema (Table \ref{tab:fem_taxonomy}) that establishes primary categories (variables, functions, operators, domains and integration variables) and determines, for each one, invariant Concept properties and instance-specific Occurrence attributes. These can be categorical (multiple choice) or single-valued (boolean, numerical, textual). On the other side, it encompasses the \textbf{compiler grammar}: We define a symbolic grammar in terms of a list of operators such as \texttt{grad()} (gradient), \texttt{div()} (divergence) or \texttt{contract()} (tensor contraction), to be utilized from the Sympy-based compiler. These will be available within the MioFFAn symbolic code synthesis palette.

This characterization is the one applied within all figures of this document, and marks the setting for the evaluations in the next section. The full files containing the complete taxonomy and operators list can be found within the examples directory of the MioFFAn repository.

\section{Automation Evaluation}
\label{sec:proof_of_concept_evaluation}

We conduct a preliminary evaluation to assess the efficacy of the MioFFAn framework and its provisional LLM routines on a curated set of variational formulations from FEM literature.

\subsection{Setup and Dataset}
\label{secsec:dataset_and_setup}

The framework was setup to interface a local vLLM server hosting a Qwen3-4B-Instruct-2507 \cite{yang_qwen3_2025} model with 65000 tokens of context length. This particular LLM model was chosen for its reported good performance in scientific content and mathematics, while being able to run on a consumer-grade GPU (RTX 4090 in our setup). A comparison against other models in the same parameters count range is shown in Appendix \ref{ap:LLM_models_comparison}.

The evaluation set consists of 6 samples sourced via the ScienceDirect API. The candidate samples collection was done in a greedy way by searching for keywords such as "finite element" or "variational" and keeping only those works published under a Creative Commons license. These were further filtered manually to ensure that they incorporate mathematical formulation aligned with the use case. Furthermore, we did a selective pruning of the original XML files, in order to remove irrelevant content and keep the HTML (in string format) within 20000 tokens long. We define the gold standard data through expert manual annotation via MioFFAn. We provide the data and processing scripts for this study within the examples directory of the MioFFAn repository.

\subsection{Evaluation Methodology}
\label{secsec:evaluation_methodology}

We evaluate the performance of the automation routines by comparing LLM-generated annotation files against their counterparts for the ground truth. To ensure valid comparisons, we setup the automated framework to perform the Concept Assignment and SoG Identification stages parting from the ground-truth outputs of the Symbol Segmentation stage; this ensures a one-to-one mapping of symbols and prevents error propagation from initial segmentation.

The following metrics are utilized within each automation phase:
\begin{itemize}
    \item Symbol Segmentation. We assess Coverage, defined as the ratio of <mi> tags correctly identified relative to the gold standard. We define a tag as correctly identified if it is assigned to an Occurrence within both the predicted and ground-truth sets, regardless of the specific placeholder Concept. This does not account for the tags that are being identified in the automated results but not in the ground truth, however we consider these cases to be a minority (most symbols should be identified in the ground truth).
    To evaluate the accuracy of symbol clustering (distribution of <mi> tags among placeholder Concepts), we utilize the standard coreference metric: CoNLL score (average of MUC, B3, and CEAFe).
    \item Concept Assignment. The CoNLL score is reapplied here to determine if the segmented symbols are mapped to the correct final Concepts.
    \item SoG Identification. To measure the quality of the Sources of Grounding (SoG), we treat the identified text and formula IDs as sets and employ the Sørensen–Dice (SDI) index. This index measures the overlap between the LLM-predicted highlights and the expert ground truth for each Occurrence. The final score for a given sample is the average of Sørensen–Dice indexes from all Occurrences.
\end{itemize}

\subsection{Results and Discussion}
\label{secsec:results_and_discussion}

In this sub-section, we present a proof-of-concept evaluation of different automation methodologies. Specifically, we compare two slightly different variants of our task automation implementations:
\begin{itemize}
    \item Original Characters (original): No modification is done to the HTML content given to the LLM. So non-standard Unicode characters will stay as they are, e.g "$\lambda$".
    \item Explicit Names (modification): The HTML given to the LLM is processed in order to replace all non-standard Unicode characters by the explicit names given by Python's \lstinline{unicodedata.name()} function, e.g. "GREEK SMALL LETTER LAMBDA".
\end{itemize}

For each approach and metric, the scores for all samples are averaged. These results are shown in Table \ref{tab:llm_results}, where we see that the Original Characters variant yields superior performance in the Symbol Segmentation task, with an advantage of over 20\% in terms of coverage and 10\% in terms of clustering accuracy. Meanwhile, the Explicit Names approach provides a marginal advantage in both Concept Assignment and SoGs Identification tasks. Such results may guide the developer towards choosing one strategy for the first task and another one for the two latter tasks. Therefore, proving the potential of the MioFFAn tool as an automation development playground.

Apart from this proof-of-concept comparison, we can take the Original Characters results as the current performance of the automation system. In this sense, the system is far from being ideal in the Symbol Segmentation task and, mostly, in the SoGs Identification task. Qualitative analysis for the latter one suggests that the model often identifies correct semantic context but struggles with the precise specification of tag IDs, sometimes highlighting entire paragraphs instead of a small sentence. These results reinforce our decision to prioritize a human-in-the-loop interface, as the automated annotations should serve only as a baseline that can be rapidly refined by a human expert, rather than a final result.

\section{MioFFAn in External Pipelines}
\label{sec:mioffan_in_external_pipelines}

We briefly discuss how MioFFAn could fit within other tools and pipelines in the field of mathematical formula processing and annotation. In the context of OCR and document mining for scientific content, standard pipelines such as Mathpix\footnote{\url{https://mathpix.com/}} and Nougat \cite{blecher_nougat_2023} focus on high-fidelity visual transcription (LaTeX or MathML) but typically lack semantic depth. MioFFAn could serve as a semantic post-processor for these outputs by appending explicit Concepts and symbolic codes to the formulas as textual metadata. Furthermore, within the Mathematical Information Retrieval field, reviewed by \citet{malik_review_2026}, MioFFAn could enhance both text-based and tree-based formula retrieval systems. One could use annotated symbolic code as a precursor for generating semantic trees, or incorporate Concepts and SoGs as additional textual information to improve formula searchability and context-aware similarity comparisons. Finally, MioFFAn may be used as an input tool for knowledge bases such as the ones presented in \cite{schembera_towards_2025} with little modification efforts, as MioFFAn’s data structure aligns closely with their framework: mathematical formulations correspond to EoIs, while quantities and their metadata map to Concepts and Concept properties.

\section{Limitations}
\label{sec:limitations}

Arguably, the biggest limitation of the MioFFAn software at current time is the sourcing system's dependency on the ScienceDirect API. Apart from integrating other API's that directly work with XML/HTML formats, conversion from PDF or LaTeX files to MioFFAn-compatible HTML format should be enabled. Annotating web content would also be challenging because of its dynamic nature. This could be solved by developing an output modality that provides the annotations in an explicit way to be appended to the web content or by embedding annotation information directly onto corresponding MathML nodes.
Other limitations include not being able to integrate links to external knowledge bases natively, not checking the applicability of CAS operators on given Concepts automatically and not allowing concurrent, collaborative annotation. All of these functionalities can be added progressively in the future.

\section{Conclusions and Future Work}
\label{sec:conclusions_and_future_work}

We have presented MioFFAn, a document-centric, interactive, and customizable software for the annotation of symbolic code corresponding to mathematical expressions. This software is proposed as a direct solution to the current scarcity of ground-truth data required for symbolic code generation in real-world research and development fields.

By extending the original MioGatto architecture, MioFFAn addresses critical challenges in structural flexibility through compound and hierarchical MI management, a decoupled properties system, and flexible source grounding. Furthermore, it incorporates the necessary features for the Formula Formalization task, specifically Equation of Interest selection and aided symbolic code specification. To demonstrate MioFFAn’s adaptability to specialized scientific domains, we have detailed a use case in Finite Element Method by configuring an appropriate concept taxonomy and operator library.

A major feature of this framework is its modular partial automation pipeline via LLMs. By decomposing parts of the annotation process into well-defined, implementation-agnostic sub-tasks, we enable researchers to iteratively test and evaluate distinct automation strategies. Our preliminary evaluation confirms that even provisional LLM routines provide a significant baseline for informed refinement, validating the effectiveness of our human-in-the-loop design.

Moving forward, we aim to enhance MioFFAn through refining current automation strategies and specifying further automation tasks, including a wider range of APIs for source gathering and support for manual document uploads, and incorporating external knowledge bases to enrich the context of identified Concepts.

\section{Acknowledgements}

Nicolas Sibuet acknowledges the Secretariat of Universities and Research of the Department of Research and Universities of the Generalitat of Catalonia, as well as the European Social Plus Fund for their financial support through the predoctoral scholarship AGAUR-FI (2024 FI-1 00089) Joan Oró.

\section{Bibliographical References}\label{sec:reference}

\bibliographystyle{lrec2026-natbib}
\bibliography{bibliography}


\appendix

\section{LLM models comparison}
\label{ap:LLM_models_comparison}

\begin{table}[!ht]
\begin{center}
\begin{tabular}{l|rr|}
\cline{2-3}
& \multicolumn{2}{l|}{Symbol Segm.} \\ \cline{2-3} 
& \multicolumn{1}{l|}{Coverage (\%)} & \multicolumn{1}{l|}{CoNLL (\%)} \\ \hline
\multicolumn{1}{|l|}{Nemotron} & \multicolumn{1}{r|}{33.9}     & \multicolumn{1}{r|}{26.4}\\ \hline
\multicolumn{1}{|l|}{Gemma} & \multicolumn{1}{r|}{28.2}     & \multicolumn{1}{r|}{18.8}\\ \hline
\multicolumn{1}{|l|}{Qwen} & \multicolumn{1}{r|}{57.7}     & \multicolumn{1}{r|}{36.8}\\ \hline
\end{tabular}
\caption{Evaluation scores for the Symbol Segmentation task, comparing three different LLM models. For each model and metric, the results from all samples have been averaged to produce a unique score.}
\label{tab:llm_results_annex}
 \end{center}
\end{table}

\begin{table}[!ht]
\begin{center}
\begin{tabular}{l|r|r|}
\cline{2-3}
& Concepts Assign. & SoGs Ident. \\ \cline{2-3} 
& CoNLL (\%) &  Avg. SDI (\%) \\ \hline
\multicolumn{1}{|l|}{Nemotron}  & 98.0  & 15.1 \\ \hline
\multicolumn{1}{|l|}{Gemma} & 95.6  & 18.3 \\ \hline
\multicolumn{1}{|l|}{Qwen} & 96.9  & 14.9 \\ \hline
\end{tabular}
\caption{Evaluation scores for the Concept Assignment and the SoG Identification tasks, comparing three different LLM models. For each model and metric, the results from all samples have been averaged to produce a unique score.}
\label{tab:llm_results_annex_2}
\end{center}
\end{table}

Three LLM models are chosen for comparison in the current implementation of the automation framework: NVIDIA's NVIDIA-Nemotron-3-Nano-4B-BF16 \cite{nvidia_nvidia_2025}, Google's gemma-3-4b-it \cite{team_gemma_2025} and Alibaba's Qwen3-4B-Instruct-2507 \cite{yang_qwen3_2025}.

The evaluation data is the same as described in Section \ref{secsec:dataset_and_setup} and the evaluation methodology is the one described in Section \ref{secsec:evaluation_methodology}.

We run each model on all samples and compute the average scores for all of them. These results are listed in Table \ref{tab:llm_results_annex} for the Symbol Segmentation task and in Table \ref{tab:llm_results_annex_2} for the Concepts Assignment and SoG Identification tasks. The performance in the two latter tasks is similar for all models, with less than 2\% of standard deviation. However, the variability of results for the Symbols Segmentation task is significant: the Qwen3-4B-Instruct-2507 model achieves an advantage of over 20\% compared to the other models in terms of coverage, and of over 10\% in terms of clustering accuracy. Given this significant difference, we choose Qwen3-4B-Instruct-2507 as the model to use in our proof-of-concept study in Section \ref{secsec:results_and_discussion}.

\end{document}